\documentclass{article}

\PassOptionsToPackage{numbers, sort&compress}{natbib}

\usepackage[preprint]{neurips_data_2021}

\usepackage[utf8]{inputenc} 
\usepackage[T1]{fontenc}    
\usepackage{hyperref}       
\usepackage{url}            
\usepackage{booktabs}       
\usepackage{amsfonts}       
\usepackage{nicefrac}       
\usepackage{microtype}      
\usepackage{xcolor}         

\usepackage{graphicx}
\usepackage{amsmath,amssymb}
\usepackage{kotex}
\usepackage[linewidth=1pt]{mdframed}

\title{Hangul Fonts Dataset: a Hierarchical and Compositional Dataset for Investigating Learned Representations}

\author{%
  Jesse A. Livezey$^{1,2}$
  \And
Ahyeon Hwang$^{3}$
\And
Jacob Yeung$^{1}$
\And
Kristofer E. Bouchard$^{1,2,4,5}$ \\
$^1$Biological Sciences and Engineering Division, Lawrence Berkeley National Laboratory, Berkeley, CA\\
$^2$Redwood Center for Theoretical Neuroscience, University of California, Berkeley, CA\\
$^3$Mathematical, Computational and Systems Biology, University of California, Irvine, CA\\
$^4$Helen Wills Neuroscience Institute, University of California, Berkeley, CA\\
$^5$Computational Research Division, Lawrence Berkeley National Laboratory\\
Berkeley, CA\\
\texttt{\{jlivezey, kebouchard\}@lbl.gov},
\texttt{ahyeon.hwang@uci.edu, jacobyeung@berkeley.edu}
}

\begin{document}

\maketitle

\begin{abstract}
Hierarchy and compositionality are common latent properties in many natural and scientific datasets. Determining when a deep network's hidden activations represent hierarchy and compositionality is important both for understanding deep representation learning and for applying deep networks in domains where interpretability is crucial. However, current benchmark machine learning datasets either have little hierarchical or compositional structure, or the structure is not known. This gap impedes precise analysis of a network's representations and thus hinders development of new methods that can learn such properties. To address this gap, we developed a new benchmark dataset with known hierarchical and compositional structure. The Hangul Fonts Dataset (HFD) is comprised of 35 fonts from the Korean writing system (Hangul), each with 11,172 blocks (syllables) composed from the product of initial consonant, medial vowel, and final consonant glyphs. All blocks can be grouped into a few geometric types which induces a hierarchy across blocks. In addition, each block is composed of individual glyphs with rotations, translations, scalings, and naturalistic style variation across fonts. We find that both shallow and deep unsupervised methods only show modest evidence of hierarchy and compositionality in their representations of the HFD compared to supervised deep networks. Supervised deep network representations contain structure related to the geometrical hierarchy of the characters, but the compositional structure of the data is not evident. Thus, HFD enables the identification of shortcomings in existing methods, a critical first step toward developing new machine learning algorithms to extract hierarchical and compositional structure in the context of naturalistic variability.
\end{abstract}

\section{Introduction}
Advances in machine learning, and representation learning in particular, have long been accompanied by the creation and detailed curation of benchmark datasets~\citep{Dua2019, lecun2010mnist, deng2009imagenet, timit_1993, wang2019glue}. Often, such datasets are created with particular structure believed to be representative of the types of structures encountered in the world. For example, many image datasets have varying degrees of hierarchy and compositionality, as exemplified by parts-based decompositions, learning compositional programs, and multi-scale representations~\citep{lee1999learning, lake2015, denton2015deep}. In contrast, synthetic image datasets often have known, (at least partial) factorial latent structure~\citep{dsprites17, Aubry14, 3dshapes18}. Having a detailed understanding of the structure of a dataset is critical to interpret the representations that are learned by any machine learning algorithm, whether linear (e.g., independent components analysis) or non-linear (e.g., deep networks). Learned representations can be used to understand the underlying structure of a dataset. Indeed, one of the desired uses of machine learning in scientific applications is to learn latent structure from complex datasets that provide insight into the data generation process~\citep{murdoch2019interpretable, raghu2020survey, aiforscience}. Understanding how learned representations relate to the structure of the training data is an area of active research~\citep{saxe2013, lake2018generalization, zeiler2014visualizing, raghu2017svcca}.

\begin{figure}[htbp!]
\centering
\includegraphics[width=5in]{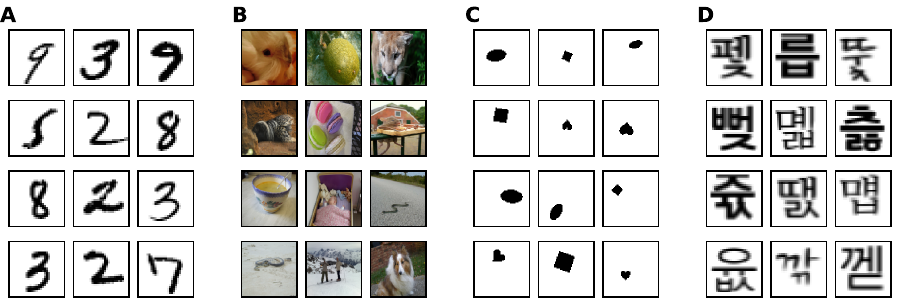}
\caption{\textbf{Ground-truth hierarchy and compositionality are lacking in benchmark machine learning datasets.} \textbf{A} Samples from the MNIST dataset. \textbf{B} Samples from the ImageNet dataset. \textbf{C} Samples from the dsprites dataset. \textbf{D} Samples from the Hangul Fonts Dataset.}\label{fig:datasets}
\end{figure}

Early benchmark image datasets such as MNIST (Fig~\ref{fig:datasets}A) and CIFAR10/100~\citep{lecun2010mnist, krizhevsky2014cifar} enabled research into early convolutional architectures. Large image datasets like ImageNet (Fig~\ref{fig:datasets}B) and COCO~\citep{deng2009imagenet, lin2014microsoft} have fueled the development of networks that can solve complex tasks like pixel-level segmentation and image captioning. Although these datasets occasionally have known semantic hierarchy (ImageNet classes are derived from the WordNet hierarchy~\citep{deng2009imagenet, miller1995wordnet}) or labeled attributes which may be part of a compositional structure (attributes like ``glasses'' or ``mustache'' in the CelebA dataset~\citep{liu2015faceattributes}), the overall complexity of these images prevents a quantitative understanding of how the hierarchy or compositionality is reflected in the data or deep network representations of the data. On the other hand, synthetic benchmark datasets such as dsprites (Fig~\ref{fig:datasets}C), and many similar variations~\citep{dsprites17, lamblin2010important, Aubry14, 3dshapes18}, have known factorial latent structure~\citep{schmidhuber1992learning}. However, these datasets typically do not have (known) hierarchy or compositionality. Thus, benchmark datasets, which have known hierarchical and compositional structure with naturalistic variability, are lacking.

A number of methods have been proposed to uncover ``disentangeled'' latent structure from images~\citep{cheung2014discovering, higgins2017beta, burgess2018understanding, chen2018isolating, rippel2014learning, schmidhuber1992learning, radford2015unsupervised, singh2018finegan, achille2018emergence, kazemi2019style, bengio2013replearn, olshausen1996emergence, bell1997independent, lee1999learning} and understand hierarchical structures in data and how they are learned in deep networks~\citep{nickel2017poincar, saxe2013}. For datasets where the form of the generative model is not known, deep representation learning methods often look for factorial or disentagled representations~\citep{cheung2014discovering, higgins2017beta, burgess2018understanding, higgins2018towards, chen2016infogan}. While factorial representations are useful for certain tasks like sampling~\citep{schmidhuber1992learning}, they do not generally capture hierarchical or compositional structures. Deep networks can learn feature hierarchies, wherein features from higher levels of the hierarchy are formed by the composition of lower level features. The hierarchical multiscale RNN captures the latent hierarchical structure by encoding the temporal dependencies with different timescales on for character-level language modelling and handwriting sequence generation tasks~\citep{chung2016hierarchical}. Deep networks have been shown to learn acoustic, articulatory, and visual hierarchies when trained on speech acoustics, neural data recorded during spoken speech syllables, and natural images, respectively~\citep{nagamine2017understanding, livezey2018deep, kell2018task, yamins2014performance}. Developing methods to probe representations for hierarchical or compositional structures is important to develop in parallel to benchmark machine learning datasets.

In this work, we present the new Hangul Fonts Dataset (HFD) (Fig~\ref{fig:datasets}D) designed for investigating hierarchy and compositionality in representation learning methods. The HFD contains a large number of data samples (391,020 across 35 fonts), annotated hierarchical and compositional structure, and naturalistic variation. Together these properties address a gap in benchmark datasets for deep learning, and representation learning research more broadly. To give examples of the potential use of the HFD, we explore whether typical deep learning methods can be used to uncover the underlying generative model of the HFD. We find that deep unsupervised networks to do recover the hierarchical or compositional latent structure, and supervised deep networks are able to partially recover the hierarchy latent structure. Thus, the Hangul Fonts Dataset will be useful for future investigations of representation learning methods.

\section {The Hangul Fonts Dataset}\label{sec:hfd}
The Korean writing system (Hangul) was created in the year 1444 to promote literacy~\cite{hangul2}. Since the Hangul writing system was partially motivated by simplicity and regularity, the rules for combining phoneme characters (and their respective glyphs) into syllables (blocks) are regular and well specified. The Hangul alphabet consists of jamo (characters) broken into 19 initial consonants, 21 medial vowels, and 27+1 final consonants (including no final consonant) which generate $19\times21\times28=11,172$ possible character blocks each of which corresponds to a syllable. Not all combinations are used in the Korean language, however all possible blocks were generated for use in this dataset. The Hangul Fonts Dataset (HFD) uses this prescribed structure as annotations for each image. The dataset consists of images of all blocks drawn in 35 different open-source fonts from~\citep{naverfonts, googlefonts, seoulfonts, iropkefonts} for a total of 391,020 annotated images.

Each Hangul block can be annotated most simply as having initial, medial, and final (IMF) independent generative variables. However, there are also variables corresponding to a geometrical hierarchy and variables corresponding to compositions of glyphs. The hierarchical variables are induced by the geometrical layout of the blocks. There are common atomic glyphs used across the initial, medial, and final characters (after a set of possible translations, rotations, and scalings)~\cite{hangul1}. The compositional variables indicate which atomic glyphs are used for each block (in a ``bag-of-atoms'' representation). Together, these different descriptions of the data facilitate investigation into what aspects of this structure representation learning methods will learn when trained on the HFD.

\subsection{The structure of a block: hierarchy and compositionality}
\begin{figure}[htbp!]
\centering
\includegraphics[width=5in]{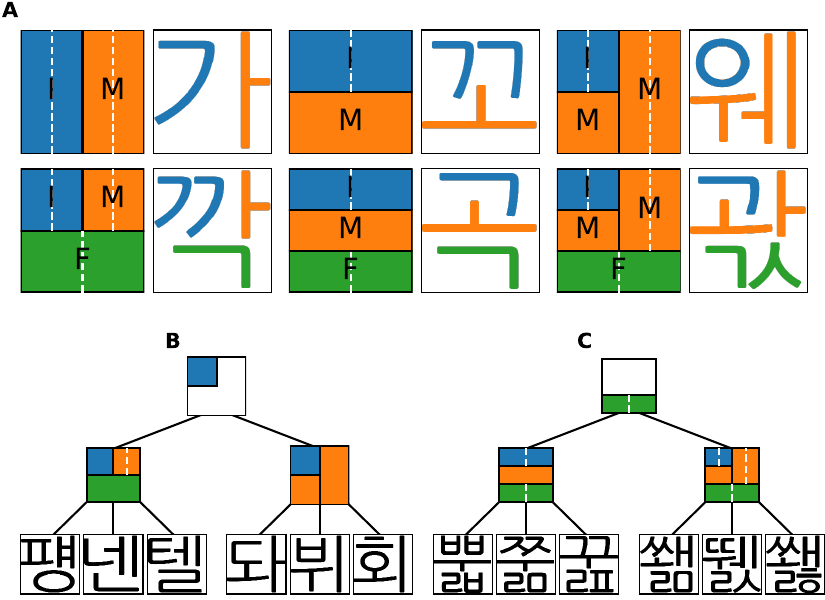}
\caption{\textbf{Hierarchy in the Hangul Fonts Dataset.} {\bf A, Hierarchy:} Each block can be grouped by the initial, medial, and/or final geometry. Block geometry and example blocks are shown. Blue indicates the possible locations of initial characters, orange indicates the possible locations of medial characters, and green indicates the possible locations of final characters. A white dashed line indicates that either a single or double character can appear. {\bf B, C, Example hierarchies:} The bottom row of the hierarchy are individual blocks. Each triplet of blocks fall under one of the geometric categories from {\bf A} (middle row) which defined the 2-level hierarchy. Then, a third level can be defined for initial, medial, or final geometric categories (top row).}\label{fig:hierarchy}
\end{figure}

There are geometrical rules for creating a block from atomic character glyphs. The initial consonant is located on the left or top of the block as either single or double characters (ㄱ and ㄲ in Fig~\ref{fig:hierarchy}A) There are 5 possible medial vowel character types: below, right-single, right-double, below-right-single, and below-right-double (ㅗ,ㅏ,ㅔ, ㅘ, and ㅞ in Fig~\ref{fig:hierarchy}A). The final consonant is at the bottom of the block as single, double, or absent characters (ㄱ and ㄳ in Fig~\ref{fig:hierarchy}A). The 30 geometrical possibilities together induce a 2-level hierarchy in the individual IMF labels and blocks that share geometric structure. The geometric variables describe the coarse layout (high level) of a block which is shared by many IMF combinations (low level) (Fig~\ref{fig:hierarchy}B and C, bottom and middle levels). Additionally, the 30 geometrical categories can be split into their intial, medial and final geometries (Fig~\ref{fig:hierarchy}B and C, bottom and middle levels). The geometric context of a character can change the style of the glyph within a block for a specific font, which is relevant for the representation analysis in Section~\ref{sec:reps}. The medial character geometry can have a large impact on how an initial glyph is translated and scaled in the block. Similarly, the final geometry can impact the scaling of the initial and medial glyphs. These contextual dependencies can be searched for in learned representations of the data. For example, a supervised deep network trained to predict the initial character may use information about the medial geometry early in the network but then eventually discard that information when predicting the initial class.

\begin{figure}[htbp!]
\centering
\includegraphics[width=5in]{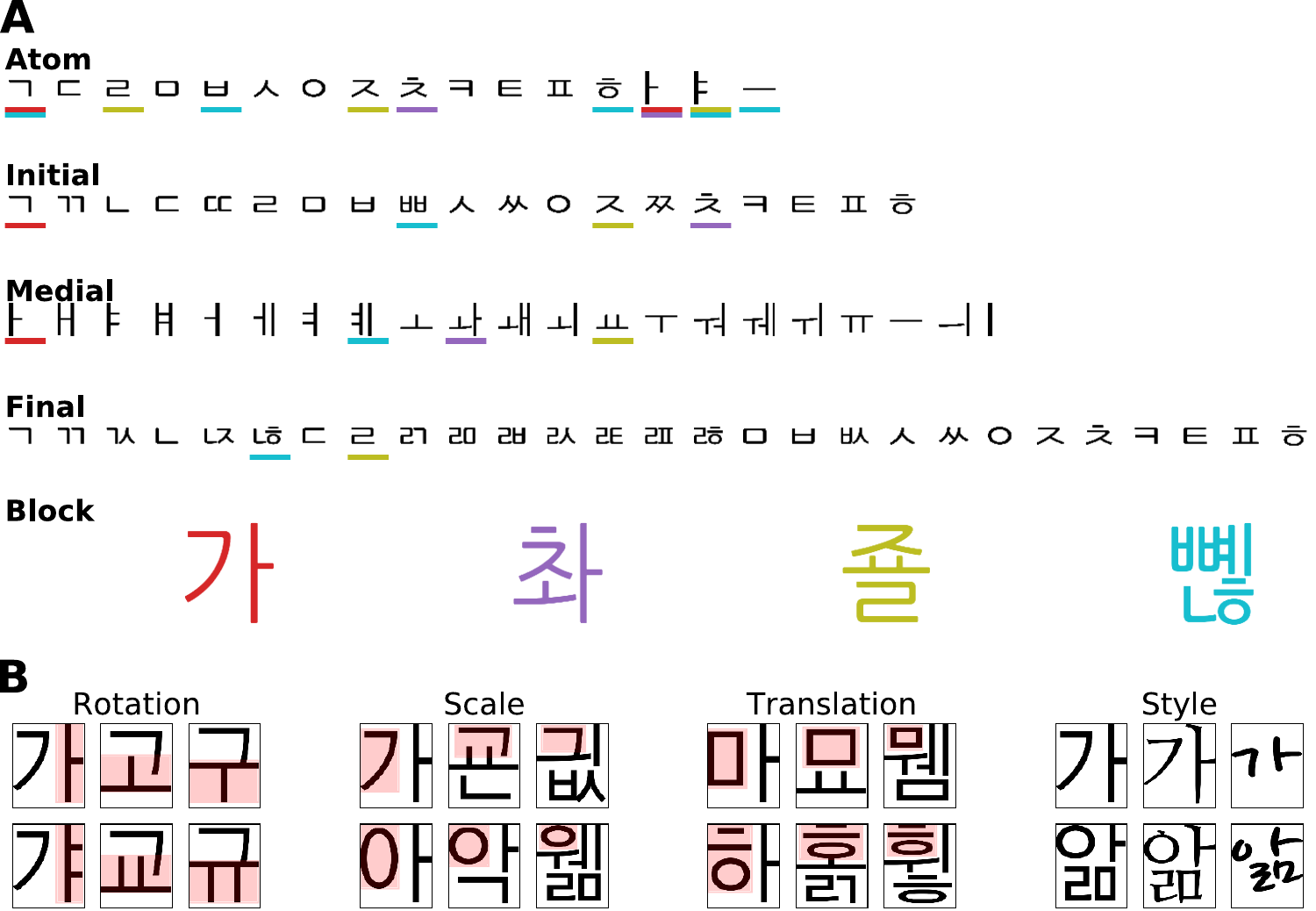}
\caption{\textbf{Composition and variation in the Hangul Fonts Dataset.} {\bf A, Composition:} Each block is composed of a set of atomic glyphs. The Atom row shows the atomic set of glyphs when scale, translations, and rotations are taken into account. The Initial, Medial, and Final (IMF) rows show all possible IMF glyphs. The Block row shows 4 example blocks with different types of structure. The color of the block is used to underline the IMF glyphs that compose the block and Atoms that compose the IMFs. {\bf B, Variability:} Two example glyphs (rows) across three different contexts (columns) are shown for each type of variation. \textbf{Rotation:} Left-most block is rotated once counterclockwise in the next block, then twice counterclockwise in the final block. \textbf{Scale:} Size of initial glyph decreases from left to right as highlighted in red. \textbf{Translation:} Highlighted glyph takes on various shapes as it is translated to different regions of the block. \textbf{Style:} Less to more stylized from left to right.}\label{fig:comp_var}
\end{figure}

Embedded within the IMF and geometrical structure is a set of glyph compositions. There are a base set of atomic glyphs from which all IMF glyphs are drawn (Fig~\ref{fig:comp_var}A, Atom row). Then, one initial, medial, and final glyph is composed into a block (Fig~\ref{fig:comp_var}A, IMF and Block rows). In this view, each block is built from a composition of potential rotations applied to a base set of atoms which are then structured by the geometrical rules and composed into a block. The underlines in the Atom and IMF rows of Fig~\ref{fig:comp_var}A correspond to inclusion in the final colored blocks in the bottom row. In this paper, for comparisons with learned representations, the composition structure is encoded in 2 ways (although the full structure is available in the dataset). The first is a ``bag-of-atoms'' binary feature set where each block is given a binary feature vector which contains a 1 if the block contains at least one atom from the top row of Fig~\ref{fig:hierarchy}A (16 features). The second is a ``bag-of-atoms'' binary feature set where the rotations have not been taken into account (24 features). These two feature sets do not encode the complete compositional structure, but they are amenable to common representation comparison methods.

The size and shape of a glyph can change within a font depending on the context. Some of these changes are consistent across fonts and stem from the changing geometry of a block with different initial, medial, or final contexts (Fig~\ref{fig:hierarchy}). Different types of variations such as rotation, translation, and more naturalistic style variations arise in the dataset (Fig~\ref{fig:comp_var}B). Glyphs can incorporate different rotations, scalings, and translation during composition into a block (Fig~\ref{fig:comp_var}B, left 3 sets). There are variations across fonts due to the nature of the design or style of the characters. These include the style of characters which can vary from clean, computer font-like fonts to highly stylized fonts which are meant to resemble hand-written characters (Fig~\ref{fig:comp_var}, rightmost set). Line thickness and the degree to which individual glyphs overlap or connect also vary. This variation is specific to a font and is based on the decision the font designer made, analogous to hand-written digits (i.e., MNIST).These types of variation are the main source of naturalistic variation in the dataset since they cannot be exactly described, but could potentially be modeled~\citep{lake2015, bell1997independent}.

\subsection{Generating the dataset}
We created a text file for the 11,172 blocks using the Unicode values from~\citep{hangulunicode}. We then converted the text files to an image file using the \texttt{convert} utility~\citep{imagemagick2008} and font files. The image sizes were different across blocks within a font, so the images were resized to the max image size across blocks. The image sizes of blocks were also different across fonts, so the blocks were resized to the median size across fonts. Individual images for the initial, medial, and final characters are included, when availble. The exact scripts used to generate the dataset, a Dockerfile which can be used to recreate or extend the HFD, and curated open fonts are provided (see Appendix A). Further summary statistics for the dataset can be found in Appendix B.

\section{Searching for hierarchy and compositionality in learned representations}\label{sec:reps}

Both shallow and deep learning models create representations (or transformations) of the input data. Methods like Principal Components Analysis (PCA) produce linear representations and Nonnegative Matrix Factorization (NMF) produces a shallow nonlinear representation inference in a linear generative model, and deep networks produce an increasingly nonlinear set of representations for each layer. Here, we compare the learned representation in unsupervised shallow methods, deep variational autoencoders, and deep feedforward classifiers. We consider whether the learned representations are organized around any of the categorical labels and hierarchy variables with an unsupervised $k$-nearest neighbor analysis. Then, we investigate whether the hierarchy or compositionality variables can be decoded with high accuracy from few features in the representations.

\begin{figure}[htbp!]
	\centering
	\includegraphics[width=5in]{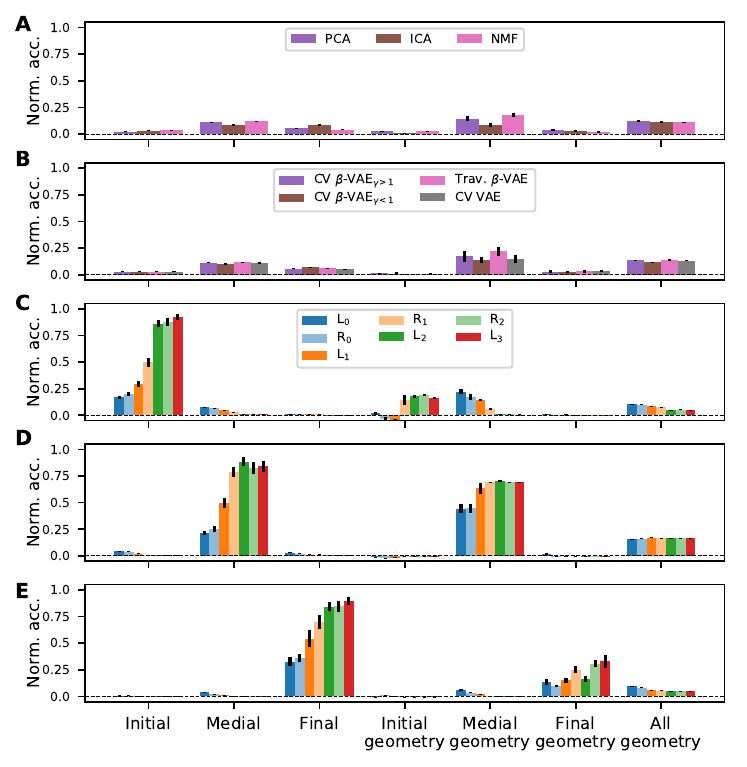}
	\caption{\textbf{Representation learning methods partially recover the geometric hierarchy.} Normalized clustering accuracy $\pm$ s.e.m. is shown across training targets, latent generative variables, layers (L is the linear part, R is after the ReLU), and model types. \textbf{A}  Normalized clustering accuracies for representations learned with unsupervised linear models. \textbf{B} Normalized clustering accuracies for representations learned with various deep VAE models. \textbf{C-E} Normalized clustering accuracies for deep representations trained to predict the initial, medial, and final label, respectively.}
	\label{fig:rep_vs_label_knn_dnn}
\end{figure}

It is desirable that deep network representations can be used to recover the generative variables of a dataset. However, it is currently not known whether deep network representations are typically organized around generative variables. In order to understand this, we test whether the latent hierarchical structure of the Hangul blocks is a major component of the learned representations using unsupervised clustering of the representations. We compare the hierarchy geometry classes from Fig~\ref{fig:hierarchy}A to $k$-nearest neighbor clusterings of the test set representations (where $k$ is set to the number of class in consideration, for more details, see Appendix C). For the shallow and deep unsupervised methods (Fig~\ref{fig:rep_vs_label_knn_dnn}A and B), we find that the medial label and geometry, final label, and all\_geometry variables are all marginally present ($0 < \text{normalized accuracy} \le 0.25$, see Section~\ref{sec:methods} for definition) in the representations. The other variables are not recovered by the unsupervised methods ($\text{normalized accuracy} \approx 0$). This shows that while VAE variants may be able to disentangle factorial structure in data, they are not well suited to extracting geometric hierarchy from the HFD with high fidelity.

In contrast (and unsurprisingly), supervised deep networks cleanly extract and recover the label they are trained on (Fig~\ref{fig:rep_vs_label_knn_dnn}C-E, first 3 colums) with increasing accuracy across layers ($\text{Norm. acc.} > 0.25$). When trained on the initial label, the initial, medial, and all\_geometry variables can all be marginally recovered, highlighting the contextual dependence of the initial glyph on the medial geometry. The initial geometry is not present in the first 2 layers, but becomes marginally present in the final layers. When trained on the medial labels, the medial geometry is present with high accuracy and the all geometries labels are marginally present. When trained on the final labels, the final geometry becomes present by the last 2 layers. There is a small amount of interaction with the medial geometry, but it is not as large as the initial-medial interaction. These results indicate that supervised deep networks do learn representations that mirror aspects of the hierarchical structure of the dataset that are most relevant for the task, and generally do not extract non-relevant hierarchy information.

\begin{figure}[htbp!]
	\centering
	\includegraphics[width=\linewidth]{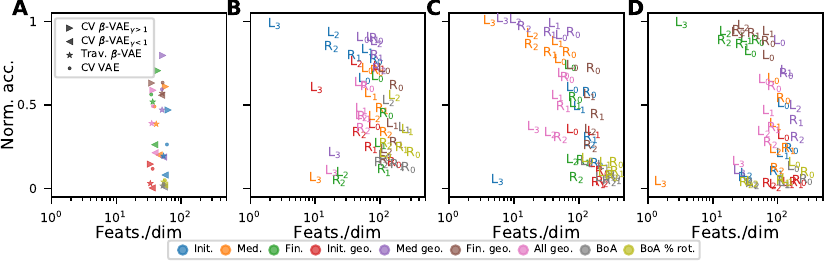}
	\caption{\textbf{Hierarchy and compositionality are not typically represented locally in deep networks.} Held-out logistic regression normalized accuracy is shown versus the ratio of the number of features selected to the variable dimensionality. Color indicates latent variable type. {\textbf A:} Results from the VAE model variants. Shape is model type. {\bf B-D:} Results from supervised deep networks trained on the initial, medial, and final tasks, respectively. Letters in correspond to the layers from Fig~\ref{fig:hierarchy}.}
	\label{fig:rep_vs_label_lr}
\end{figure}

Understanding whether deep network representations tend to be more distributed or local is an open area of research~\citep{zeiler2014visualizing, nguyen2016multifaceted, olah2018the}. We investigated whether deep networks learn a local representation by training sparse logistic regression models to predict the latent hierarchy and compositionality variables from the representations (Fig~\ref{fig:rep_vs_label_lr}). If the representation of a hierarchy or compositionality variable is present and simple (linear), we would expect the normalized accuracy to be high (near 1 on the y-axis of the plots in Fig~\ref{fig:rep_vs_label_lr}). If a representation of a variable is ``local'', we would expect the varible to be decoded using approximately the same number of features as it has dimensions (near $10^1$ on the x-axis of Fig~\ref{fig:rep_vs_label_lr}) and ``distributed'' representation to have a much higher ratio. To test this, we compare these two measures across models and target variables and also across layers for the supervised deep networks.

We find that unsupervised ($\beta$-)VAEs (Fig~\ref{fig:rep_vs_label_lr}A) learn consistently distributed representations of the latent variables (typically 30-60x more features than the variable dimension are selected). In terms of the prediction accuracy, the cross validated $\beta$-VAEs tend to have higher accuracy across variables than the VAE and the $\beta$-VAE selected for traversals, although there is a fair amount of heterogeneity. For supervised deep networks (Fig~\ref{fig:rep_vs_label_lr}B-D), the supervision variable (initial, medial, final, respectively), has high accuracy across layers, and moves from a more distributed to a more local representation at deep layers. For the initial and medial labels, the medial geometry can also be read out with high accuracy and an increase in localization across layers. The initial geometry is not read out with high accuracy in the initial and medial label networks, and the final geometry variable can only be predicted well for the final label network. The all\_geometry variable can be predicted at marginal accuracy for all networks. The compositional Bag-of-Atoms (BoA) features cannot be predicted well (often at or below chance) for any network and the BoA mod rotations can only be read out with marginal accuracy for the initial label network. These results suggest that standard, fully-connected deep networks do not typically learn local representations for variables except for those they are trained on (and correlated variables).

\section{Methods}\label{sec:methods}
\subsection{Representation learning methods}
Principal Component Analysis (PCA), Independent Component Analysis (ICA), and Non-negative Matrix Factorization (NMF) from Scikit-Learn~\citep{pedregosa2011scikit} were used to learn representations from the data. These methods were all trained with 100 components which is at least 3-times larger than any of the latent generative variables under consideration. The models were trained on the training and validation sets and the representation analysis was on the test set.

\begin{figure}
    \centering
    \includegraphics[width=5in]{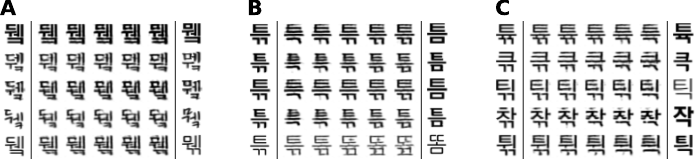}
    \caption{\textbf{Disentangled reconstructions from $\beta$-VAE.} Latent traversals of a single latent variable. The left column is the input image, middle columns are the traversals, and right column is the block the traversals appear to morph into. \textbf{A, Initial Across Fonts:} First four rows are similar traversals of an initial glyph from one block across increasingly naturalistic fonts.  Final row is an entangled traversal between initial and final glyphs. \textbf{B, Initial Across Blocks:} First four rows are similar traversals of a final glyph from one block across different fonts. Final row is an entangled traversal between initial, medial, and final glyphs. \textbf{C, Final Across Fonts:} First two rows are similar traversals of a final glyph from blocks (with the same hierarchy) in the same font. Third row is a traversal of a final glyph from a block (with a different hierarchy). Fourth row is an entangled traversal between initial and final glyphs. Final row shows an entangled traversal of medial and final glyphs.}
    \label{fig:traversals}
\end{figure}

Variational autoencoders (VAEs) learn a latent probabalistic model of the data they are trained on. The $\beta$-VAE is a variant of a VAE which aims to learn disentangled latent factors~\citep{higgins2017beta, burgess2018understanding} by trading off the reconstruction and KL-divergence terms with a factor different than 1. We implement the $\beta$-VAE from \citet{burgess2018understanding}, which encourages the latent codes to have a specific capacity. We experiment with both $\beta > 1$ from \citep{burgess2018understanding} as well as $\beta < 1$ from \citep{Alemi2018FixingAB, Snderby2016LadderVA}.  $\beta$-VAE networks with convolutional and dense layers were trained on the dataset. 100 sets of hyperparameters were used for training the $\beta$-VAEs. The hyperparameters and their ranges are listed in Appendix D. In order to cross-validate the networks, we checked if the same blocks across fonts are nearest neighbors in the latent space. For each block in each font, the nearest neighbor is found. If the neighbor has the same label as the block, we assign an accuracy of 1, otherwise 0. This is averaged across all blocks and pairs of fonts in the validation set. The model with the best cross-validation accuracy for each label was chosen and the downstream analysis was done on the test set latent encodings. We also cherry-picked networks which had interpretable latent traversals (Fig~\ref{fig:traversals}).

Fully-connected networks with 3 hidden layers were trained on one of the initial, medial, or final glyph variables. For each task, 100 sets of hyperparameters were used for training. The hyperparameters and their ranges are listen in Appendix D. The model with the best validation accuracy was chosen and the downstream analysis was done on the test set representations (test accuracies reported in Appendix B). Code for training the networks and reproducing the figures will be posted publicly. Deep networks representation analysis was partially completed on the NERSC supercomputer. All deep learning models were trained using PyTorch~\citep{paszke2017pytorch} on Nvidia GTX 1080s or Titan Xs.

To compare accuracies (and chance accuracies) across models with differing numbers of classes (between 2 and 30), we 0-1 normalize the accuracies across models to make comparisons more clear. Specifically, for a model with given $\text{accuracy}=a$ and $\text{chance}=c$, we report
\begin{equation}
    \text{Norm. acc.} = \frac{a - c}{1 - c}
\end{equation}
which is 0 when $a=c$ 1 if $a=1$, independent of the number or distribution of classes.

\subsection{Generative structure recovery from representation of the data}
The 35 fonts were used in a 7-fold cross validation loop for the machine learning methods. The fonts were randomly permuted and then 5 fonts were used for each of the non-overlapping validation and test sets. The analysis of representations was done on the test set representations. For the supervised deep networks, the following methods were applied to the activations of every layer both before and after the ReLU nonlinearities. For the unsupervised VAEs, they were applied to samples from the latent layer. The logistic regression analysis was not applied to the linear representations.

Clustering a representation produces a reduced representation for every datapoint in an unsupervised way. If one chooses the number of clusters to be equal to the dimensionality or number of classes the generative variables has, then they can be directly compared (up to a permutation). We cluster the representations with KMeans and then find the optimal alignment of the real and clustered labels (see Appendix C for more details). We then report the normalized accuracy of this labeling across training variables, layers, and hierarchy variables.

Sparse logistic regression attempts to localize the information about a predicted label into a potentially small set of features. To do this, we used logistic regression models fit using the Union of Intersection (UoI) method~\citep{bouchard2017union, sachdeva2019pyuoi}. The UoI method has been shown to be able to fit highly sparse models without a loss in predictive performance~\citep{sachdeva2021improved}. We report the normalized accuracy and mean number of features selected divided by the number of features or classes across training variables, layers, and hierarchy variables. For this analysis, 2 new training and testing sub-splits were created from the representations on the original test set that was held out during deep network training.

\section{Discussion}\label{sec:discussion}
The Hangul Fonts Dataset (HFD) presented here has hierarchical and compositional latent structure that allows each image (block) to have ground-truth annotations, making the HFD well suited for deep representation research. Using a set of unsupervised and supervised methods, we are able to extract a subset of the variables from the representations of deep networks. Several VAE variants have relatively poor variable recovery from their latent layers, while supervised deep networks have clear representation of the variables they are trained on and interacting variables. Understanding how to better recover such structure from deep network representations will broaden the application of deep learning in science.

In many scientific domains like cosmology, neuroscience, and climate science, deep learning is being used to make high accuracy predictions given growing dataset sizes~\citep{mathuriya2018cosmoflow, livezey2018deep, kim2019deep, livezey2021deep}. However, deep learning is not commonly used to directly test hypotheses about dataset structure. This is partially because the nonlinear, compositional structure of deep networks, which is conducive to high accuracy prediction from complex data, is not ideal for interrogating hypotheses about data. In particular, it is not generally known how the structure of a dataset influences the learned data representations or whether the structure of the dataset can be ``read-out'' of the learned representations. Understanding which dataset structures can be extracted from learned deep representations is important for the expanded use of deep learning in scientific applications.

The HFD is based on a set of fonts which provide some naturalistic variation. However, the amount of variation is likely much smaller than what would be found in a handwritten dataset of Hangul characters. One benefit to using fonts is that the dataset can be easily extended as new fonts are created. To this end, we release the entire dataset creation pipeline to aid in future expansion of the HFD or the creation of similar font-based datasets. Another potential limitation and area of future work is determining how to encode variables like hierarchy and compositionality. In this dataset, there is a natural class-based encoding for the shallow geometry hierarchy. The Bag-of-Atoms composition encoding ignores structure that is potentially relevant for recovering compositionality (much like Bag-of-Words features discard potentially useful structure in natural language processing).

In this work, relatively small fully-connected and convolutional networks were considered. However, these techniques can be applied to larger feedforward networks, recurrent networks, or networks with residual layers to understand the impact on learned representations. Understanding how proposed methods for learning factorial or disentangled representations~\citep{schmidhuber1992learning, cheung2014discovering, higgins2017beta, achille2018emergence} impact the structure of learned representations is important for using deep network representations for hypothesis testing in scientific domains. Compared to disentangling~\citep{higgins2018towards}, relatively little work addresses how to define and evaluate hierarchy and compositionality in learned representations. Furthermore, unsupervised or semi-supervised cross-validation metrics that can be used for model selection across a range of structure recovery tasks (e.g., disentangling, hierarchy recovery, compositionality recovery) are lacking.

\begin{ack}
JAL, AH, and KEB were supported by the Deep Learning for Science Laboratory Directed Research and Development. We are grateful for the feedback on the project from the Neural Systems and Data Science Lab.
\end{ack}

\vfill
\pagebreak
\appendix

\section{Generating the Hangul Fonts Dataset}
To increase the reproducibility and potential for future development of the Hangul Fonts Dataset (HFD), we are releasing the code used to generate and normalize the HFD. The code is currently undergoing our institutional review process for licensing, but will be released under a \href{https://spdx.org/licenses/BSD-3-Clause-LBNL.html}{Lawrence Berkeley National Labs BSD variant license} by the time of publication along with a DOI from a corresponding Zenodo archive. The code and instructions for reproducing the HFD, including a docker image and Dockerfile which specifies the environment, can be found at \url{https://github.com/BouchardLab/HangulFontsDatasetGenerator}. The 35 open font files, their licences, and a number of additional fonts have been curated \href{https://drive.google.com/file/d/1AgabLzjGX_tDrA0mKvYhJGuGmWYWD8RW/view?usp=sharing}{here}.

This code (with default settings) generates the following
\begin{itemize}
    \item \texttt{pngs/*:} folders per font and fontsize (default 24) containing png files of each block
    \item \texttt{pdfs/*:} folder per font containing pdfs which show all possible blocks
    \item \texttt{h5s/*:} folder per font containing HDF5 files which contain the following (key: variable) pairs
    \begin{itemize}
        \item \texttt{images:} block images at max size across blocks within the font (not used directly)
        \item \texttt{images\_median\_shape:} block images at median size across fonts (used for training/analysis)
        \item \texttt{labels:} concatenated IMF class labels
        \item \texttt{\{initial, medial, final\}\_geometry:} IMF geometry hierarchy class labels
        \item \texttt{all\_geometry:} class labels based on the product of the IMF geometry class labels
        \item \texttt{atom\_bof:} IMF atoms features
        \item \texttt{atom\_mod\_rotations\_bof:} IMF atoms mod rotations features
    \end{itemize}
\end{itemize}

\section{The Hangul fonts dataset: summary statistics and visualization}
\subsection{Font families}
\begin{figure}[htbp!]
\centering
\includegraphics[width=1\textwidth]{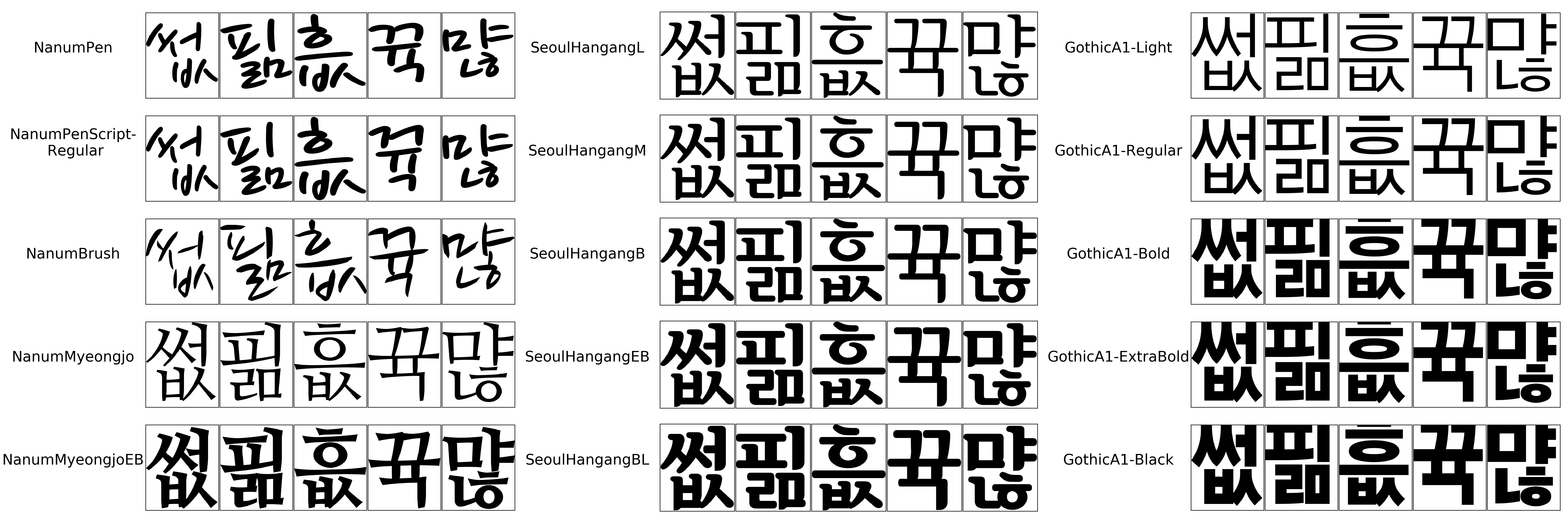}
\caption{\textbf{Variation across different font families: Nanum, SeoulHangang, GothicA1.} The same five randomly chosen blocks are displayed for each font. SeoulHangang and GothicA1 fonts are ordered thin to bold from top to bottom.}\label{font_comparison}
\end{figure}

For certain fonts, bold and light versions of the same font are included, a ``natural'' but fairly regular source of variation across fonts (Fig~\ref{font_comparison}).

\begin{figure}[htbp!]
\centering
\includegraphics[width=1\textwidth]{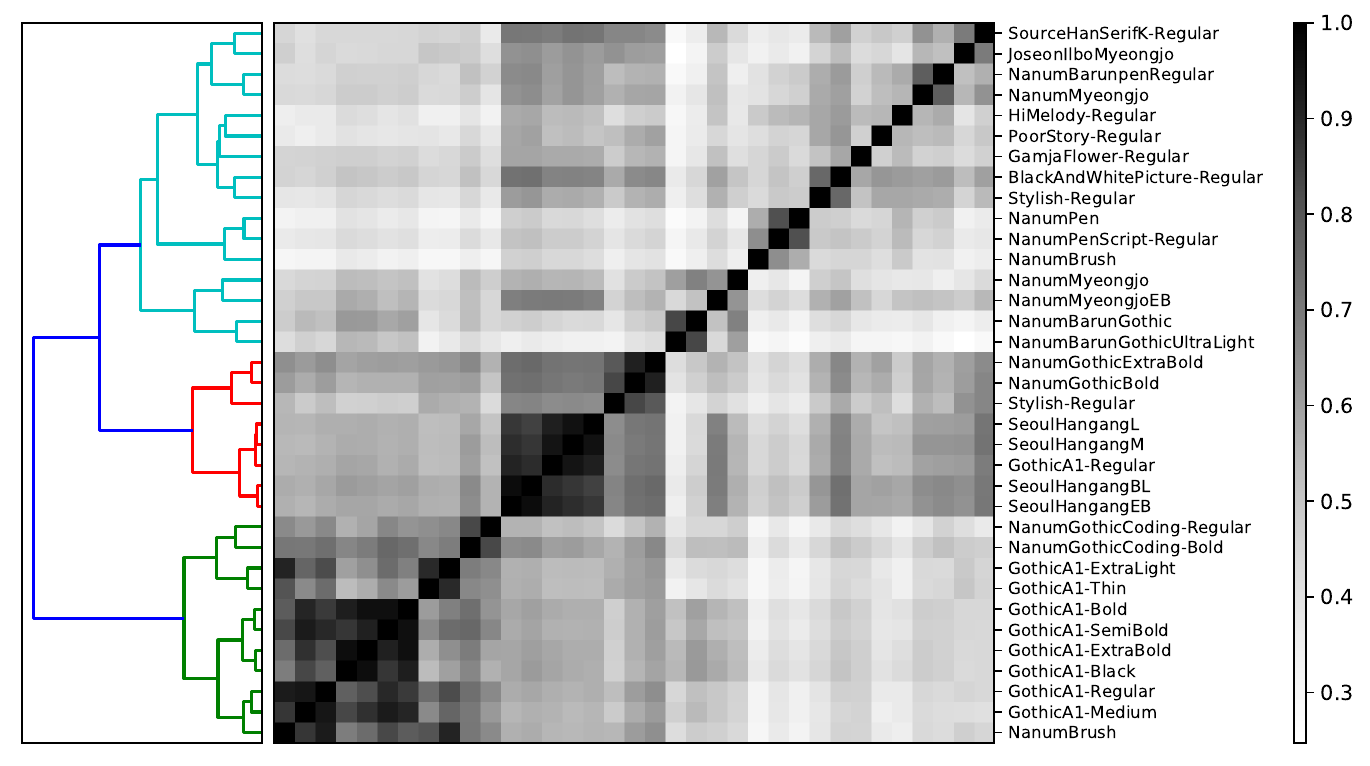}
\caption{\textbf{Distance matrix and dendrogram showing clustering of similar fonts.} Distinct groups such as Nanum, SeoulHangang, and GothicA1 cluster together.}\label{dendrogram}
\end{figure}

We also performed correlations across fonts to analyze how fonts differ from one another. Fig~\ref{dendrogram} shows the dendrogram that results from hierarchical clustering of the correlation matrix using Ward's method. GothicA1-Bold, GothicA1-SemiBold, GothicA1-Regular, GothicA1-Light, GothicA1-Thin, GothicA1-Black, GothicA1-Medium, and GothicA1-ExtraBold all fall under the same GothicA1 category.

\subsection{Summary statistics of dataset}
\begin{figure}[htbp!]
\centering
\includegraphics[width=0.5\textwidth]{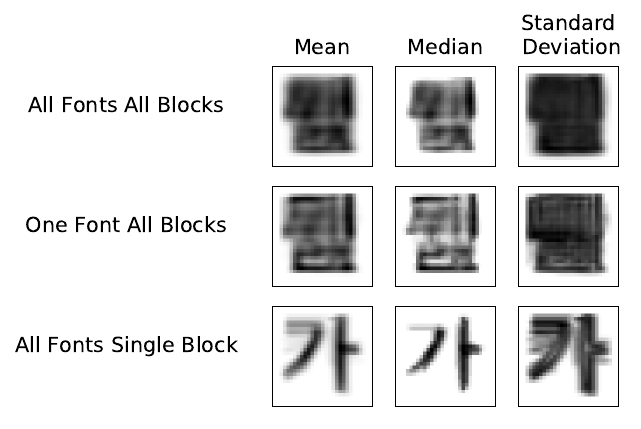}
\caption{\textbf{Summary statistics of images in Hangul dataset.} The mean, median, and standard deviation is taken for all blocks in all fonts, all blocks in a single font, and a single block '가' in all fonts.}\label{mms}
\end{figure}

\begin{figure}[htbp!]
\centering
\includegraphics[width=1\textwidth]{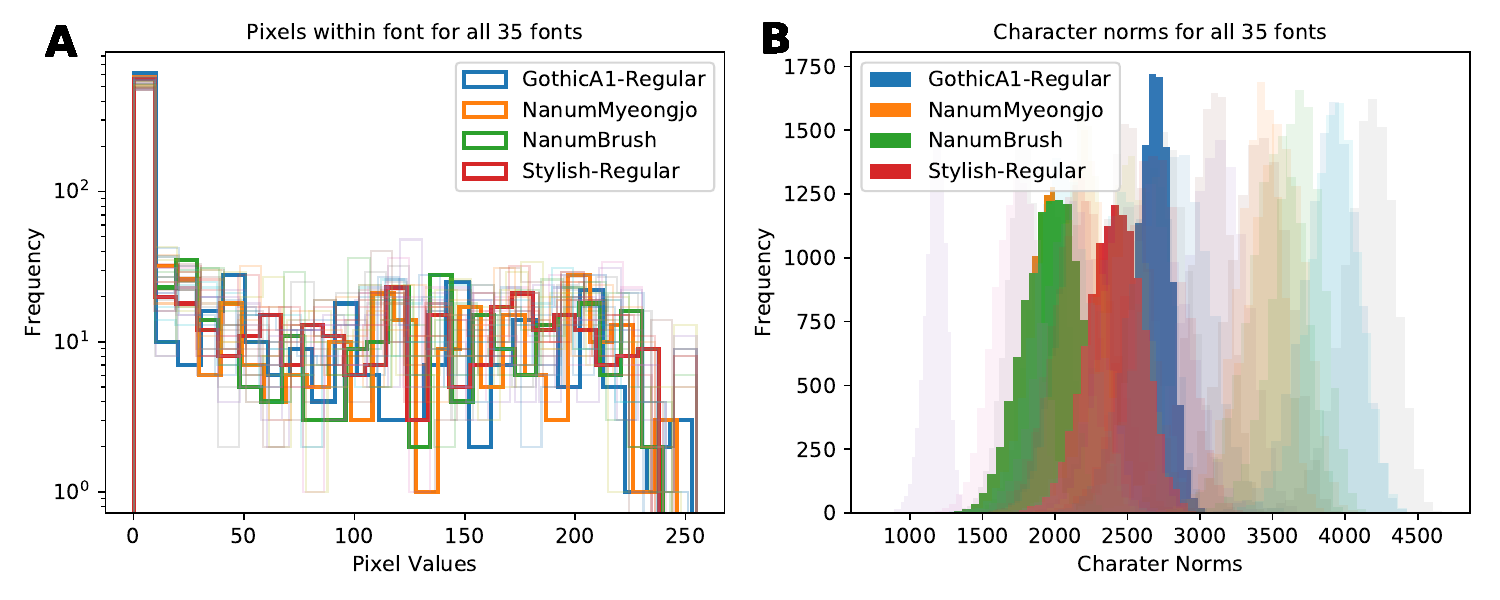}
\caption{\textbf{Characteristics of pixels and characters for each font.} Four fonts are shown clearly while the rest are shown in a lighter shade. \textbf{A} Outlines of the histograms of pixels within a font. \textbf{B} Filled-in histograms of pixels within font across all blocks.}\label{hist}
\end{figure}

Various statistics were calculated on fontsize 24 images within a single and across all fonts. The mean, median, and standard deviation of the images were taken in Fig~\ref{mms}. This was done for all blocks in all fonts, all blocks in a single font, and a single block in all fonts. All three statistics for all fonts all blocks and one font all blocks preserve the block structure of the images, whereas '가' is clearly shown for all fonts single block across the statistics. Fig~\ref{hist}A shows a histogram of the pixels of all the images within a font for all 35 fonts. The histograms are very similar across the four fonts highlighted in the legend. Fig~\ref{hist}B shows a histogram of pixels within a font across all characters for all 35 fonts. The Frobenius norm is taken for all characters to study how characters differ within a font. NanumMyeongjo and NanumBrush are similar fonts as they have overlapping character norms. GothicA1-Regular, which resembles computer-type fonts, has the thinnest distribution as its characters do not differ greatly.

\subsection{Dimensionality Reduction: PCA, ICA, NMF}
\begin{figure}[htbp!]
\centering
\includegraphics[width=1\textwidth]{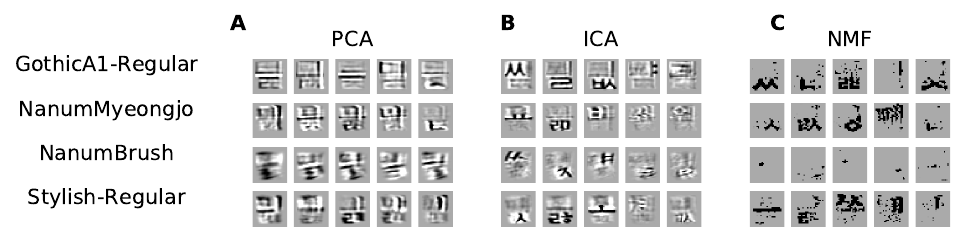}
\caption{\textbf{Dictionary elements learned from four fonts: GothicA1-Regular, NanumMyeongjo, NanumBrush, Stylish-Regular.} \textbf{A} PCA elements: first five PCs that explain the most variance are plotted from left to right. \textbf{B} ICA elements: first 3 ICs show distinct glyphs, last 2 are randomly chosen ICs \textbf{C} NMF elements: first three best factors are chosen, last 2 are randomly chosen. NanumBrush had no distinct components.}
\label{fig:rep}
\end{figure}

Linear models were trained on individual fonts and the learned dictionaries are shown in Fig~\ref{fig:rep}.

\subsection{Classification accuracy}
The supervised deep networks were trained on 3 tasks: initial, medial, and final (IMF) classification. Here we report the test-set accuracy for logistic regression as well as the best deep networks (selected by the validation accuracy). Logistic regression had accuracies of $57.6\pm2.7\%$, $70.8\pm2.6\%$, and $73.0\pm3.1\%$ respectively for the IMF tasks. Deep networks had accuracies of $98.5\pm2.0\%$, $98.0\pm2.6\%$, and $97.4\pm3.3\%$ respectively for the IMF tasks. Chance accuracy is $5.3\%$, $2.6\%$, and $3.1\%$ respectively for the IMF tasks.

\subsection{UMAP visualization}
\begin{figure}[htbp!]
\centering
\includegraphics[width=1\textwidth]{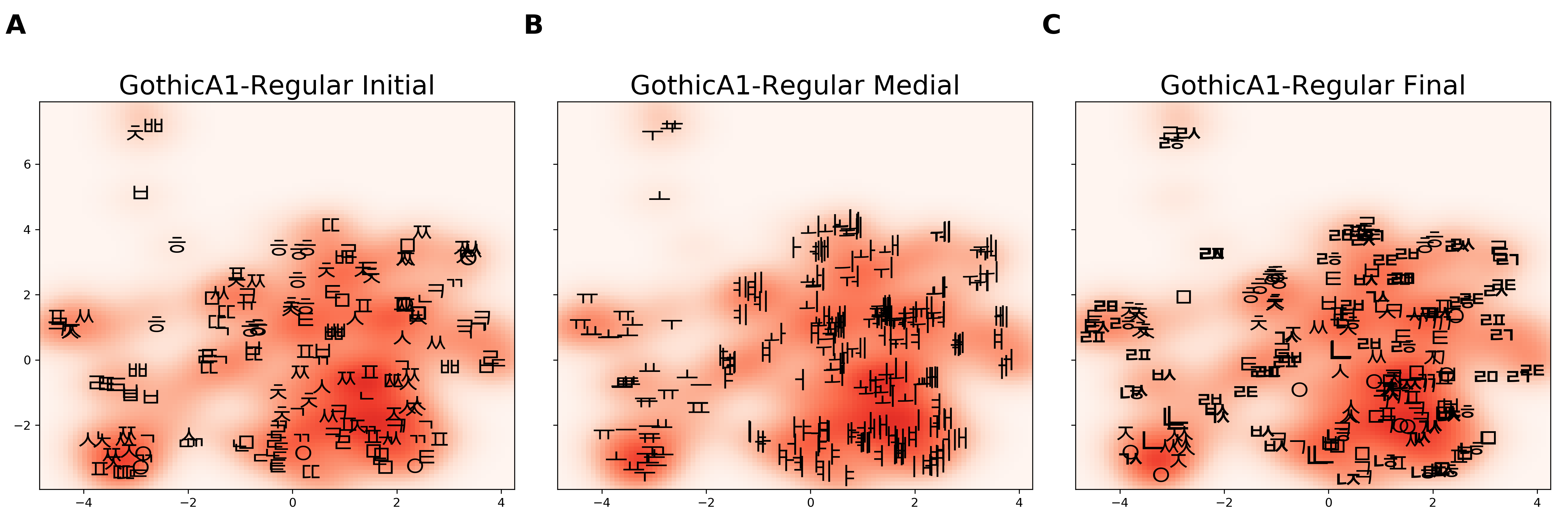}
\caption{\textbf{GothicA1-Regular UMAP embedding of Initial, Medial, and Final Labels.} The kernel density plot indicated by the red coloration is the same for all three figures. Darker shade means higher density of characters. \textbf{A} 19 initial glyphs, 10 of each plotted \textbf{B} 21 medial glyphs, 10 of each plotted \textbf{C} 28 final glyphs, 8 of each plotted  }
\label{fig:umap_chars}
\end{figure}

Fig~\ref{fig:umap_chars}A-C shows the result of applying UMAP to a single font's images, GothicA1-Regular, with initial, medial, and final labels. Individual glyphs are plotted with a red kernel density estimate in the background. Fig~\ref{fig:umap_chars}B shows the best clustering with regards to the geometry of the glyphs. The more verticially-oriented glyphs (ㅏ,ㅑ,ㅕ,ㅓ) cluster together in the right side while the more horizontally-oriented glyphs (ㅛ,ㅡ,ㅜ,ㅠ) cluster in the left. In Fig~\ref{fig:umap_chars}C several of the duplet glyphs embed in the same location, suggesting similarity in the duplet structure.

\begin{figure}[htbp!]
\centering
\includegraphics[width=1\textwidth]{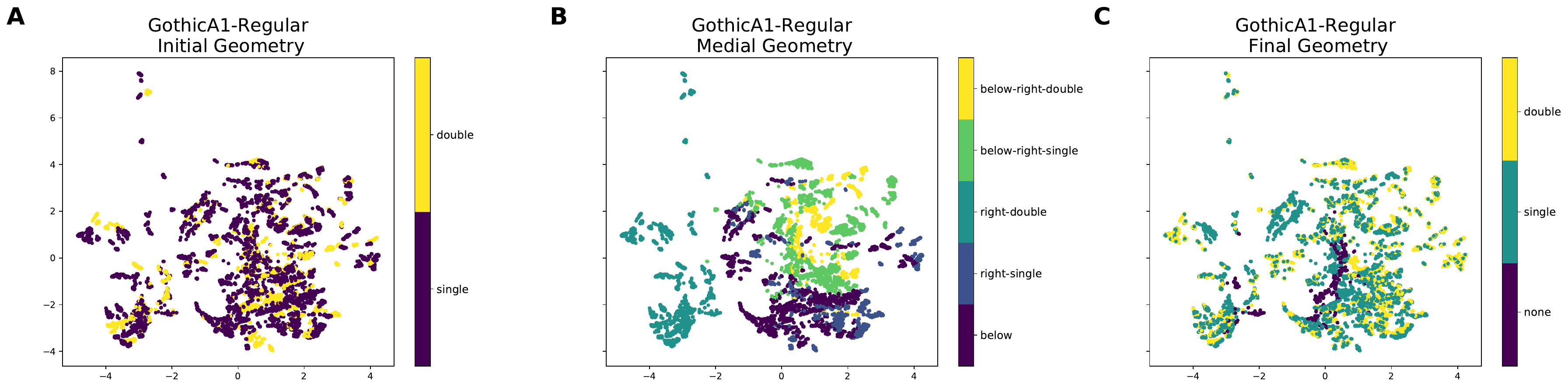}
\caption{\textbf{GothicA1-Regular UMAP embedding of Initial, Medial, and Final Geometry Labels.} \textbf{A} 2 initial geometric types \textbf{B} 5 medial geometric types \textbf{C} 3 final geometric types}
\label{fig:umap_geom}
\end{figure}

Fig~\ref{fig:umap_geom}A-C shows the result of applying UMAP to the images with initial, medial, and final geometry labels. Actual points are plotted rather than glyphs. Fig~\ref{fig:umap_geom}B shows the best separation among the 5 geometric types as they are each distinct, and hence affect the overall structure of the block character. For example, right-single and right-double medial geometric types are always placed on the left region of the blocks. In contrast, initial and final geometry types which include none, single, or double do not drastically influence the greater structure of the block. Single and double geometric types have very similar embeddings in both the initial and final geometry plots.

\section{Methods continued}
\subsection{KMeans clustering accuracy}
For a dataset of sample size $n$ and a representation $h\in \mathbb{R}^{n\times d}$ we want to compare a clustering of $h$ and a categorical generative variable $y\in \{1,\ldots,M\}$. First, $h$ is clustered into $M$ clusters with k-means (with k=$M$) or from cutting the dendrogram to give $M$ clusters. Given this clustering, each sample is assigned a class label $\hat p$. We then have to find the best alignment between the $M$ cluster labels and the $M$ generative variable labels. To do this we form a similarity matrix $S\in \mathbb{R}^{M\times M}$. To calculate $S_{ij}$, we first form the set $A$ which contains the samples labeled $\hat p_i$ and $B$ which contains the samples labeled $y_j$. Then the similarity is the cardinality of the intersection of the sets divided by the cardinality of the union of the sets: $S_{ij}=\tfrac{|A\cap B|}{|A \cup B|}$, that is, given all samples labeled with either label, what fraction of them are labelled as both. We then use the Hungarian method~\cite{kuhn1955hungarian} to optimally pair the generative labels $y$ with a permutation of the cluster labels $\hat y$ using this similarity matrix. If the cluster labelling is an exact permutation, the clustering accuracy will be 100\%, and chance for a random relabelling.

\section{Hyperparameters}
All supervised networks were trained with 3 hidden layers of the same dimensionality and ReLU nonlinearities. Table~\ref{table:hyp} lists the hyperparameters that were randomly sampled and their ranges. $\dim_i$ and $\dim_0$ indicate the input and output dimensionality of the data and task.

Unsuperised $\beta-$VAE encoders were trained with 7 $2-$d convolutional layers of the same channel depth and kernel size, ReLU nonlinearities, and one hidden layer of the same dimensionality. The decoders were trained with one hidden layer of the same dimensionality, transposed $2-$d convolutional layers of the same channel depth and kernel size, and ReLU nonlinearities. The $\gamma < 1$ networks were trained with the training regime from \cite{burgess2018understanding}. The $\gamma > 1$ networks were trained with the same training regime as $\gamma < 1$ for 125 epochs (using $\gamma < 1$), after which the $\gamma$ term was increased exponentially, similar to the training regime proposed by \cite{Alemi2018FixingAB}.

\begin{table}[h]
	\caption{Hyperparameters for dense networks}
	\label{table:hyp}
	\begin{center}
		\begin{tabular}{lll}
			\multicolumn{1}{c}{\textbf{Name}}  &\multicolumn{1}{c}{\textbf{Type}} &\multicolumn{1}{c}{\textbf{Range/Options}}
			\\ \hline \\
			Init. momentum & Float & .5 \\
			Learning rate reduction on plateau & float & .5\\
			Epochs of patience for learning rate reduction & int & 10\\
			Epochs of patience for early stopping & int & 10\\
			\\ \hline \\
			Dense layer size & int & $\min(\dim_i-1,\ \dim_o \times 10)$ : $2\times\ \dim_i$\\
			$\log_{10}$ learning rate & float & -6 : 1 \\
			$\log_{10}$(1-momentum) & float & -2 : -.00436 (momentum=.01) \\
			$\log_{10}\ L_2$ weight decay & float & -6 : 1 \\
			Batch size & int & 32 : 512 \\
			Input dropout rate & float & .1 : .99 \\
			Input dropout rescale & float & .1 : 10 \\
			Hidden dropout rate & float & .1 : .99 \\
			Hidden dropout rescale & float & .1 : 10 \\
		\end{tabular}
	\end{center}
	
	\caption{Hyperparameters for $\beta$-VAE networks}
	\label{table:vae_hyp}
	\begin{center}
	    \begin{tabular}{lll}
	        \multicolumn{1}{c}{\textbf{Name}}  &\multicolumn{1}{c}{\textbf{Type}} &\multicolumn{1}{c}{\textbf{Range/Options}}
			\\ \hline \\
			Init. momentum & Float & .5 \\
			Learning rate reduction on plateau & float & .5\\
			Epochs of patience for learning rate reduction & int & 5\\
			Epochs of patience for early stopping & int & 125\\
			\\ \hline \\
			$\log_{10}$ learning rate & float & -5 : -2 \\
			$\log_{10}$(1-momentum) & float & -2 : -.00436 (momentum=.01) \\
			$\log_{10}\ L_2$ weight decay & float & -6 : 1 \\
			$\log_{10}$ $\gamma (< 1)$ & float & -4 : -1\\
			$\log_{10}$ $\gamma (> 1)$ & float & 1 : 3\\
			$\log$ KL-divergence target (C) max value & float & 20 : 50\\
	    \end{tabular}
	\end{center}
\end{table}

\newpage

\section{Datasheet for the Hangul Fonts Dataset}

We follow the Datasheets for Datasets~\cite{gebru2018datasheets} recommendations for documenting datasets.

\subsection{Motivation}
\begin{mdframed}
For what purpose was the dataset created?\\
Who created the dataset and on behalf of which entity?\\
Who funded the creation of the dataset?
\end{mdframed}

The Hangul Fonts Dataset was created as a benchmark machine learning dataset for investigating whether and how hierarchy and compositionality are found representation learning methods with a focus on deep networks. It was created by Jesse Livezey, Ahyeon Hwang, and Kristofer Bouchard at Lawrence Berkeley National Laboratory (LBNL) and was funded through the Laboratory Directed Research and Development Program at LBNL.

\subsection{Composition}
\begin{mdframed}
What do the instances that comprise the dataset represent? Does the dataset contain all possible instances or is it a sample of instances from a larger set? What data does each instance consist of?\\
How many instances are there in total?\\
Is there a label or target associated with each instance?\\
Are there recommended data splits?\\
Are there any errors, sources of noise, or redundancies in the dataset?\\
Is the dataset self-contained, or does it link to or otherwise rely on external resources?\\
Does the dataset contain data that might be considered confidential? Does the dataset contain data that, if viewed directly, might be offensive, insulting, threatening, or might otherwise cause anxiety? Does the dataset relate to people?
\end{mdframed}

The Hangul Fonts Datasets consists of all possible blocks for each font along with a set of labels and features for each block which are common across all fonts. We provide scripts to render each block and font into an image and store them together in an HDF5 file for each font and fontsize specified. In addition, we have curated a set of Hangul fonts which are licensed under the TODO.

Each font has 11,172 possible blocks and we have curated 35 fonts for a total of 391,020 possible images. The spoken consonants and vowel correspond to a set of ``initial'', ``medial'', and ``final'' labels for each block with 19, 21, and 28 classes respectively. In addition, each block can also be assigned a hierarchy label based on the block geometry. Finally, the individual glyphs (atoms) are composed across the initial, medial, and final locations along with potential scalings, translations, or rotations. We use these relationship to define a bag-of-atoms set of features.

Rendering the fonts to images transforms them from a vector-based representation to a pixel-based representation which will introduce some smoothing or loss of information depending on the font size used. We find that there is not much gain in classification accuracy beyond font size 24 which is the font size we recommend and generate by default.

The dataset relies on a set of open font files, which we have curated. Otherwise, the dataset is self contained and we provide scripts for users to generate their own copy of the dataset. Using the scripts, users are also able to generate datasets with varying fontsizes or new open or closed fonts.

The dataset does not contain any data that might be considered confidential or harmful. Although by their nature fonts and the Hangul writing system relate to people in their creation and use, the instances in this dataset are not related to data from any people.

\subsection{Collection Process}
\begin{mdframed}
How was the data associated with each instance acquired? What mechanisms or procedures were used to collect the data?\\
If the dataset is a sample from a larger set, what was the sampling strategy?\\
Who was involved in the data collection process and how were they compensated?\\
Over what timeframe was the data collected?\\
Were any ethical review processes conducted?\\
\end{mdframed}
The open font files which are used in conjunction with the library to generate the images were curated from lists of Hangul fonts at TODO and TODO. All availble, open, de-duplicated fonts were used in the final dataset. Jesse Livezey and Ahyeon Hwang curated the font files and wrote the library and were paid as part of a postdoc and a research assistant position, respectively, at LBNL. The fonts were curated and the library was written between October 2017 and February 2019. No ethical review process was conducted.

\subsection{Preprocessing/cleaning/labeling}
\begin{mdframed}
Was any preprocessing/cleaning/labeling of the data done?\\
Was the “raw” data saved in addition to the preprocessed/cleaned/labeled data?\\
Is the software used to preprocess/clean/label the instances available?\\
\end{mdframed}
The vector font files are converted into a set of images. There are choices in terms of font size and final image standardization made and documented in the library. The initial, medial, and final labels, the geometry labels, and the atom labels are based on the structure of the Hangul writing system. The identification of different atoms across rotations was done by hand. In lieu of releasing the binary image files, we provide the library used to create the images along with a Docker image which can be used to exactly replicate the images.

\subsection{Uses}
\begin{mdframed}
Has the dataset been used for any tasks already?\\
Is there a repository that links to any or all papers or systems that use the dataset?\\
What (other) tasks could the dataset be used for?\\
Is there anything about the composition of the dataset or the way it was collected and preprocessed/cleaned/labeled that might impact future uses?\\
Are there tasks for which the dataset should not be used?\\
\end{mdframed}
The dataset has not been used outside of its original preprint/publication. The dataset could be used for other types of generative and supervised machine learning analysis of Hangul fonts. There were preprocessing choices made which may impact future use, but the library used to generate the images is being released. We cannot think of potential tasks for which the dataset should not be used.

\subsection{Distribution}
\begin{mdframed}
Will the dataset be distributed to third parties outside of the entity on behalf of which the dataset was created?\\
How will the dataset will be distributed (e.g., tarball on website,API, GitHub)?\\
When will the dataset be distributed?\\
Will the dataset be distributed under a copyright or other intellectual property (IP) license, and/or under applicable terms of use(ToU)?\\
Have any third parties imposed IP-based or other restrictions on the data associated with the instances?\\
Do any export controls or other regulatory restrictions apply to the dataset or to individual instances?\\
\end{mdframed}
The library used to generate the images from the font files will be distributed on Github under an open source license. We have also curated a set of open font files. The library will be distributed under a \href{https://spdx.org/licenses/BSD-3-Clause-LBNL.html}{Lawrence Berkeley National Labs BSD variant license} license after institutional review is complete. There are no restrictions or controls on the dataset or instances.

\subsection{Maintenance}
\begin{mdframed}
Who is supporting/hosting/maintaining the dataset?\\
How can the owner/curator/manager of the dataset be contacted?\\
Is there an erratum?\\
Will the dataset be updated?\\
If the dataset relates to people, are there applicable limits on the retention of the data associated with the instances?\\
Will older versions of the dataset continue to be supported/hosted/maintained?\\
If others want to extend/augment/build on/contribute to thedataset, is there a mechanism for them to do so?\\
\end{mdframed}
The dataset will be hosted on Github (with an archive or the publication version on Zenodo) and maintained by authors who can be contacted through Github and/or their institutional email addresses. The dataset will be updated to correct any errors which will be noted in a erratum. Future versions of the dataset will be extended to work with additional open fonts as they become available and contributions/corrections from others. Contributions/corrections can be made through Github. Old versions will be tagged and maintained as applicable. The dataset does not relate to people.

\small

\bibliographystyle{unsrtnat}
\bibliography{references}

\begin{thebibliography}{66}
\providecommand{\natexlab}[1]{#1}
\providecommand{\url}[1]{\texttt{#1}}
\expandafter\ifx\csname urlstyle\endcsname\relax
  \providecommand{\doi}[1]{doi: #1}\else
  \providecommand{\doi}{doi: \begingroup \urlstyle{rm}\Url}\fi

\bibitem[Dua and Graff(2017)]{Dua2019}
Dheeru Dua and Casey Graff.
\newblock {UCI} machine learning repository, 2017.
\newblock URL \url{http://archive.ics.uci.edu/ml}.

\bibitem[LeCun et~al.()LeCun, Cortes, and Burges]{lecun2010mnist}
Yann LeCun, Corinna Cortes, and CJ~Burges.
\newblock Mnist handwritten digit database.
\newblock \emph{AT\&T Labs [Online]. Available:
  http://yann.lecun.com/exdb/mnist}.

\bibitem[Deng et~al.(2009)Deng, Dong, Socher, Li, Li, and
  Fei-Fei]{deng2009imagenet}
Jia Deng, Wei Dong, Richard Socher, Li-Jia Li, Kai Li, and Li~Fei-Fei.
\newblock Imagenet: A large-scale hierarchical image database.
\newblock In \emph{2009 IEEE conference on computer vision and pattern
  recognition}, pages 248--255. Ieee, 2009.

\bibitem[Garofolo et~al.(1993)Garofolo, Lamel, Fisher, Fiscus, Pallett,
  Dahlgren, and Zue]{timit_1993}
John Garofolo, Lori Lamel, William Fisher, Jonathan Fiscus, David Pallett,
  Nancy Dahlgren, and Victor Zue.
\newblock {TIMIT Acoustic-Phonetic Continuous Speech Corpus}, 1993.
\newblock URL \url{https://hdl.handle.net/11272.1/AB2/SWVENO}.

\bibitem[Wang et~al.(2019)Wang, Singh, Michael, Hill, Levy, and
  Bowman]{wang2019glue}
Alex Wang, Amanpreet Singh, Julian Michael, Felix Hill, Omer Levy, and
  Samuel~R. Bowman.
\newblock {GLUE}: A multi-task benchmark and analysis platform for natural
  language understanding.
\newblock 2019.
\newblock In the Proceedings of ICLR.

\bibitem[Lee and Seung(1999)]{lee1999learning}
Daniel~D Lee and H~Sebastian Seung.
\newblock Learning the parts of objects by non-negative matrix factorization.
\newblock \emph{Nature}, 401\penalty0 (6755):\penalty0 788--791, 1999.

\bibitem[Lake et~al.(2015)Lake, Salakhutdinov, and Tenenbaum]{lake2015}
Brenden~M Lake, Ruslan Salakhutdinov, and Joshua~B Tenenbaum.
\newblock Human-level concept learning through probabilistic program induction.
\newblock \emph{Science}, 350\penalty0 (6266):\penalty0 1332--1338, 2015.

\bibitem[Denton et~al.(2015)Denton, Chintala, Szlam, and
  Fergus]{denton2015deep}
Emily Denton, Soumith Chintala, Arthur Szlam, and Rob Fergus.
\newblock Deep generative image models using a laplacian pyramid of adversarial
  networks.
\newblock \emph{arXiv preprint arXiv:1506.05751}, 2015.

\bibitem[Matthey et~al.(2017)Matthey, Higgins, Hassabis, and
  Lerchner]{dsprites17}
Loic Matthey, Irina Higgins, Demis Hassabis, and Alexander Lerchner.
\newblock dsprites: Disentanglement testing sprites dataset.
\newblock https://github.com/deepmind/dsprites-dataset/, 2017.

\bibitem[Aubry et~al.(2014)Aubry, Maturana, Efros, Russell, and Sivic]{Aubry14}
Mathieu Aubry, Daniel Maturana, Alexei Efros, Bryan Russell, and Josef Sivic.
\newblock Seeing 3d chairs: exemplar part-based 2d-3d alignment using a large
  dataset of cad models.
\newblock In \emph{CVPR}, 2014.

\bibitem[Burgess and Kim(2018)]{3dshapes18}
Chris Burgess and Hyunjik Kim.
\newblock 3d shapes dataset.
\newblock https://github.com/deepmind/3dshapes-dataset/, 2018.

\bibitem[Murdoch et~al.(2019)Murdoch, Singh, Kumbier, Abbasi-Asl, and
  Yu]{murdoch2019interpretable}
W~James Murdoch, Chandan Singh, Karl Kumbier, Reza Abbasi-Asl, and Bin Yu.
\newblock Interpretable machine learning: definitions, methods, and
  applications.
\newblock \emph{arXiv preprint arXiv:1901.04592}, 2019.

\bibitem[Raghu and Schmidt(2020)]{raghu2020survey}
Maithra Raghu and Eric Schmidt.
\newblock A survey of deep learning for scientific discovery.
\newblock \emph{arXiv preprint arXiv:2003.11755}, 2020.

\bibitem[Stevens et~al.(2020)Stevens, Taylor, Nichols, Maccabe, Yelick, and
  Brown]{aiforscience}
Rick Stevens, Valerie Taylor, Jeff Nichols, Arthur~Barney Maccabe, Katherine
  Yelick, and David Brown.
\newblock {AI} for science.
\newblock 2020.

\bibitem[Saxe et~al.(2013)Saxe, McClelland, and Ganguli]{saxe2013}
Andrew~M Saxe, James~L McClelland, and Surya Ganguli.
\newblock Exact solutions to the nonlinear dynamics of learning in deep linear
  neural networks.
\newblock \emph{arXiv preprint arXiv:1312.6120}, 2013.

\bibitem[Lake and Baroni(2018)]{lake2018generalization}
Brenden Lake and Marco Baroni.
\newblock Generalization without systematicity: On the compositional skills of
  sequence-to-sequence recurrent networks.
\newblock In \emph{International Conference on Machine Learning}, pages
  2873--2882. PMLR, 2018.

\bibitem[Zeiler and Fergus(2014)]{zeiler2014visualizing}
Matthew~D Zeiler and Rob Fergus.
\newblock Visualizing and understanding convolutional networks.
\newblock In \emph{European conference on computer vision}, pages 818--833.
  Springer, 2014.

\bibitem[Raghu et~al.(2017)Raghu, Gilmer, Yosinski, and
  Sohl-Dickstein]{raghu2017svcca}
Maithra Raghu, Justin Gilmer, Jason Yosinski, and Jascha Sohl-Dickstein.
\newblock {SVCCA}: Singular vector canonical correlation analysis for deep
  learning dynamics and interpretability.
\newblock \emph{arXiv preprint arXiv:1706.05806}, 2017.

\bibitem[Krizhevsky et~al.(2014)Krizhevsky, Nair, and
  Hinton]{krizhevsky2014cifar}
Alex Krizhevsky, Vinod Nair, and Geoffrey Hinton.
\newblock The cifar-10 dataset.
\newblock \emph{online: http://www. cs. toronto. edu/kriz/cifar. html}, 55,
  2014.

\bibitem[Lin et~al.(2014)Lin, Maire, Belongie, Hays, Perona, Ramanan,
  Doll{\'a}r, and Zitnick]{lin2014microsoft}
Tsung-Yi Lin, Michael Maire, Serge Belongie, James Hays, Pietro Perona, Deva
  Ramanan, Piotr Doll{\'a}r, and C~Lawrence Zitnick.
\newblock Microsoft coco: Common objects in context.
\newblock In \emph{European conference on computer vision}, pages 740--755.
  Springer, 2014.

\bibitem[Miller(1995)]{miller1995wordnet}
George~A Miller.
\newblock Wordnet: a lexical database for english.
\newblock \emph{Communications of the ACM}, 38\penalty0 (11):\penalty0 39--41,
  1995.

\bibitem[Liu et~al.(2015)Liu, Luo, Wang, and Tang]{liu2015faceattributes}
Ziwei Liu, Ping Luo, Xiaogang Wang, and Xiaoou Tang.
\newblock Deep learning face attributes in the wild.
\newblock In \emph{Proceedings of International Conference on Computer Vision
  (ICCV)}, December 2015.

\bibitem[Lamblin and Bengio(2010)]{lamblin2010important}
Pascal Lamblin and Yoshua Bengio.
\newblock Important gains from supervised fine-tuning of deep architectures on
  large labeled sets.
\newblock In \emph{NIPS* 2010 Deep Learning and Unsupervised Feature Learning
  Workshop}, pages 1--8, 2010.

\bibitem[Schmidhuber(1992)]{schmidhuber1992learning}
J{\"u}rgen Schmidhuber.
\newblock Learning factorial codes by predictability minimization.
\newblock \emph{Neural Computation}, 4\penalty0 (6):\penalty0 863--879, 1992.

\bibitem[Cheung et~al.(2014)Cheung, Livezey, Bansal, and
  Olshausen]{cheung2014discovering}
Brian Cheung, Jesse~A Livezey, Arjun~K Bansal, and Bruno~A Olshausen.
\newblock Discovering hidden factors of variation in deep networks.
\newblock \emph{arXiv preprint arXiv:1412.6583}, 2014.

\bibitem[Higgins et~al.(2017)Higgins, Matthey, Pal, Burgess, Glorot, Botvinick,
  Mohamed, and Lerchner]{higgins2017beta}
Irina Higgins, Loic Matthey, Arka Pal, Christopher Burgess, Xavier Glorot,
  Matthew Botvinick, Shakir Mohamed, and Alexander Lerchner.
\newblock beta-vae: Learning basic visual concepts with a constrained
  variational framework.
\newblock In \emph{International Conference on Learning Representations},
  volume~3, 2017.

\bibitem[Burgess et~al.(2018)Burgess, Higgins, Pal, Matthey, Watters,
  Desjardins, and Lerchner]{burgess2018understanding}
Christopher~P Burgess, Irina Higgins, Arka Pal, Loic Matthey, Nick Watters,
  Guillaume Desjardins, and Alexander Lerchner.
\newblock Understanding disentangling in $\beta$-vae.
\newblock \emph{arXiv preprint arXiv:1804.03599}, 2018.

\bibitem[Chen et~al.(2018)Chen, Li, Grosse, and Duvenaud]{chen2018isolating}
Ricky~TQ Chen, Xuechen Li, Roger Grosse, and David Duvenaud.
\newblock Isolating sources of disentanglement in vaes.
\newblock In \emph{Proceedings of the 32nd International Conference on Neural
  Information Processing Systems}, pages 2615--2625, 2018.

\bibitem[Rippel et~al.(2014)Rippel, Gelbart, and Adams]{rippel2014learning}
Oren Rippel, Michael Gelbart, and Ryan Adams.
\newblock Learning ordered representations with nested dropout.
\newblock In \emph{International Conference on Machine Learning}, pages
  1746--1754. PMLR, 2014.

\bibitem[Radford et~al.(2015)Radford, Metz, and
  Chintala]{radford2015unsupervised}
Alec Radford, Luke Metz, and Soumith Chintala.
\newblock Unsupervised representation learning with deep convolutional
  generative adversarial networks.
\newblock \emph{arXiv preprint arXiv:1511.06434}, 2015.

\bibitem[Singh et~al.(2018)Singh, Ojha, and Lee]{singh2018finegan}
Krishna~Kumar Singh, Utkarsh Ojha, and Yong~Jae Lee.
\newblock Finegan: Unsupervised hierarchical disentanglement for fine-grained
  object generation and discovery.
\newblock \emph{arXiv preprint arXiv:1811.11155}, 2018.

\bibitem[Achille and Soatto(2018)]{achille2018emergence}
Alessandro Achille and Stefano Soatto.
\newblock Emergence of invariance and disentanglement in deep representations.
\newblock \emph{The Journal of Machine Learning Research}, 19\penalty0
  (1):\penalty0 1947--1980, 2018.

\bibitem[Kazemi et~al.(2019)Kazemi, Iranmanesh, and Nasrabadi]{kazemi2019style}
Hadi Kazemi, Seyed~Mehdi Iranmanesh, and Nasser Nasrabadi.
\newblock Style and content disentanglement in generative adversarial networks.
\newblock In \emph{2019 IEEE Winter Conference on Applications of Computer
  Vision (WACV)}, pages 848--856. IEEE, 2019.

\bibitem[Bengio et~al.(2013)Bengio, Courville, and Vincent]{bengio2013replearn}
Yoshua Bengio, Aaron Courville, and Pascal Vincent.
\newblock Representation learning: A review and new perspectives.
\newblock \emph{IEEE Transactions on Pattern Analysis and Machine
  Intelligence}, 35\penalty0 (8):\penalty0 1798--1828, 2013.
\newblock \doi{10.1109/TPAMI.2013.50}.

\bibitem[Olshausen and Field(1996)]{olshausen1996emergence}
Bruno~A Olshausen and David~J Field.
\newblock Emergence of simple-cell receptive field properties by learning a
  sparse code for natural images.
\newblock \emph{Nature}, 381\penalty0 (6583):\penalty0 607--609, 1996.

\bibitem[Bell and Sejnowski(1997)]{bell1997independent}
Anthony~J Bell and Terrence~J Sejnowski.
\newblock The “independent components” of natural scenes are edge filters.
\newblock \emph{Vision research}, 37\penalty0 (23):\penalty0 3327--3338, 1997.

\bibitem[Nickel and Kiela(2017)]{nickel2017poincar}
Maximilian Nickel and Douwe Kiela.
\newblock Poincar\'e embeddings for learning hierarchical representations.
\newblock \emph{arXiv preprint arXiv:1705.08039}, 2017.

\bibitem[Higgins et~al.(2018)Higgins, Amos, Pfau, Racaniere, Matthey, Rezende,
  and Lerchner]{higgins2018towards}
Irina Higgins, David Amos, David Pfau, Sebastien Racaniere, Loic Matthey,
  Danilo Rezende, and Alexander Lerchner.
\newblock Towards a definition of disentangled representations.
\newblock \emph{arXiv preprint arXiv:1812.02230}, 2018.

\bibitem[Chen et~al.(2016)Chen, Duan, Houthooft, Schulman, Sutskever, and
  Abbeel]{chen2016infogan}
Xi~Chen, Yan Duan, Rein Houthooft, John Schulman, Ilya Sutskever, and Pieter
  Abbeel.
\newblock Infogan: Interpretable representation learning by information
  maximizing generative adversarial nets.
\newblock \emph{arXiv preprint arXiv:1606.03657}, 2016.

\bibitem[Chung et~al.(2016)Chung, Ahn, and Bengio]{chung2016hierarchical}
Junyoung Chung, Sungjin Ahn, and Yoshua Bengio.
\newblock Hierarchical multiscale recurrent neural networks.
\newblock \emph{arXiv preprint arXiv:1609.01704}, 2016.

\bibitem[Nagamine and Mesgarani(2017)]{nagamine2017understanding}
Tasha Nagamine and Nima Mesgarani.
\newblock Understanding the representation and computation of multilayer
  perceptrons: A case study in speech recognition.
\newblock In \emph{Proceedings of the 34th International Conference on Machine
  Learning-Volume 70}, pages 2564--2573. JMLR. org, 2017.

\bibitem[Livezey et~al.(2018)Livezey, Bouchard, and Chang]{livezey2018deep}
Jesse~A Livezey, Kristofer~E Bouchard, and Edward~F Chang.
\newblock Deep learning as a tool for neural data analysis: speech
  classification and cross-frequency coupling in human sensorimotor cortex.
\newblock \emph{arXiv preprint arXiv:1803.09807}, 2018.

\bibitem[Kell et~al.(2018)Kell, Yamins, Shook, Norman-Haignere, and
  McDermott]{kell2018task}
Alexander~JE Kell, Daniel~LK Yamins, Erica~N Shook, Sam~V Norman-Haignere, and
  Josh~H McDermott.
\newblock A task-optimized neural network replicates human auditory behavior,
  predicts brain responses, and reveals a cortical processing hierarchy.
\newblock \emph{Neuron}, 98\penalty0 (3):\penalty0 630--644, 2018.

\bibitem[Yamins et~al.(2014)Yamins, Hong, Cadieu, Solomon, Seibert, and
  DiCarlo]{yamins2014performance}
Daniel~LK Yamins, Ha~Hong, Charles~F Cadieu, Ethan~A Solomon, Darren Seibert,
  and James~J DiCarlo.
\newblock Performance-optimized hierarchical models predict neural responses in
  higher visual cortex.
\newblock \emph{Proceedings of the national academy of sciences}, 111\penalty0
  (23):\penalty0 8619--8624, 2014.

\bibitem[of~Korean~Language(2008{\natexlab{a}})]{hangul2}
National~Institure of~Korean~Language.
\newblock Want to know about {H}angeul?, Jan. 2008{\natexlab{a}}.
\newblock URL
  \url{https://web.archive.org/web/20190111001341/http://www.korean.go.kr/eng_hangeul/setting/002.html}.

\bibitem[Software()]{naverfonts}
Naver Software.
\newblock Naver software hangul font collections.
\newblock URL
  \url{https://software.naver.com/software/fontList.nhn?categoryId=I0000000}.

\bibitem[Files()]{googlefonts}
Google~Fonts Files.
\newblock Google fonts files.
\newblock URL \url{https://github.com/google/fontsl}.

\bibitem[Symbols()]{seoulfonts}
Seoul’s Symbols.
\newblock Seoul’s symbols.
\newblock URL
  \url{http://english.seoul.go.kr/seoul-views/seoul-symbols/5-fonts/}.

\bibitem[iropke()]{iropkefonts}
iropke.
\newblock iropke.
\newblock URL \url{http://font.iropke.com/batang/}.

\bibitem[of~Korean~Language(2008{\natexlab{b}})]{hangul1}
National~Institure of~Korean~Language.
\newblock Want to know about {H}angeul?, Jan. 2008{\natexlab{b}}.
\newblock URL
  \url{https://web.archive.org/web/20190111001835/http://www.korean.go.kr/eng_hangeul/principle/001.html}.

\bibitem[in~Korean()]{hangulunicode}
Programming in~Korean.
\newblock Hangul in unicode.
\newblock URL
  \url{https://web.archive.org/web/20190513221943/http://www.programminginkorean.com/programming/hangul-in-unicode/}.

\bibitem[ImageMagick~Studio(2008)]{imagemagick2008}
LLC ImageMagick~Studio.
\newblock Imagemagick, 2008.

\bibitem[Nguyen et~al.(2016)Nguyen, Yosinski, and
  Clune]{nguyen2016multifaceted}
Anh Nguyen, Jason Yosinski, and Jeff Clune.
\newblock Multifaceted feature visualization: Uncovering the different types of
  features learned by each neuron in deep neural networks.
\newblock \emph{arXiv preprint arXiv:1602.03616}, 2016.

\bibitem[Olah et~al.(2018)Olah, Satyanarayan, Johnson, Carter, Schubert, Ye,
  and Mordvintsev]{olah2018the}
Chris Olah, Arvind Satyanarayan, Ian Johnson, Shan Carter, Ludwig Schubert,
  Katherine Ye, and Alexander Mordvintsev.
\newblock The building blocks of interpretability.
\newblock \emph{Distill}, 2018.
\newblock \doi{10.23915/distill.00010}.
\newblock https://distill.pub/2018/building-blocks.

\bibitem[Pedregosa et~al.(2011)Pedregosa, Varoquaux, Gramfort, Michel, Thirion,
  Grisel, Blondel, Prettenhofer, Weiss, Dubourg, et~al.]{pedregosa2011scikit}
Fabian Pedregosa, Ga{\"e}l Varoquaux, Alexandre Gramfort, Vincent Michel,
  Bertrand Thirion, Olivier Grisel, Mathieu Blondel, Peter Prettenhofer, Ron
  Weiss, Vincent Dubourg, et~al.
\newblock Scikit-learn: Machine learning in python.
\newblock \emph{Journal of machine learning research}, 12\penalty0
  (Oct):\penalty0 2825--2830, 2011.

\bibitem[Alemi et~al.(2018)Alemi, Poole, Fischer, Dillon, Saurous, and
  Murphy]{Alemi2018FixingAB}
Alexander~Amir Alemi, Ben Poole, Ian~S. Fischer, Joshua~V. Dillon, R.~Saurous,
  and K.~Murphy.
\newblock Fixing a broken elbo.
\newblock In \emph{ICML}, 2018.

\bibitem[S{\o}nderby et~al.(2016)S{\o}nderby, Raiko, Maal{\o}e, S{\o}nderby,
  and Winther]{Snderby2016LadderVA}
C.~S{\o}nderby, T.~Raiko, Lars Maal{\o}e, S{\o}ren~Kaae S{\o}nderby, and
  O.~Winther.
\newblock Ladder variational autoencoders.
\newblock In \emph{NIPS}, 2016.

\bibitem[Paszke et~al.(2017)Paszke, Gross, Chintala, and
  Chanan]{paszke2017pytorch}
Adam Paszke, Sam Gross, Soumith Chintala, and Gregory Chanan.
\newblock Pytorch: Tensors and dynamic neural networks in python with strong
  gpu acceleration.
\newblock \emph{PyTorch: Tensors and dynamic neural networks in Python with
  strong GPU acceleration}, 6, 2017.

\bibitem[Bouchard et~al.(2017)Bouchard, Bujan, Roosta-Khorasani, Ubaru,
  Prabhat, Snijders, Mao, Chang, Mahoney, and Bhattacharya]{bouchard2017union}
Kristofer Bouchard, Alejandro Bujan, Farbod Roosta-Khorasani, Shashanka Ubaru,
  Mr~Prabhat, Antoine Snijders, Jian-Hua Mao, Edward Chang, Michael~W Mahoney,
  and Sharmodeep Bhattacharya.
\newblock Union of intersections ({UoI}) for interpretable data driven
  discovery and prediction.
\newblock In \emph{Advances in Neural Information Processing Systems}, pages
  1078--1086, 2017.

\bibitem[Sachdeva et~al.(2019)Sachdeva, Livezey, Tritt, and
  Bouchard]{sachdeva2019pyuoi}
Pratik~S Sachdeva, Jesse~A Livezey, Andrew~J Tritt, and Kristofer~E Bouchard.
\newblock Pyuoi: The union of intersections framework in python.
\newblock \emph{Journal of Open Source Software}, 4\penalty0 (44):\penalty0
  1799, 2019.

\bibitem[Sachdeva et~al.(2021)Sachdeva, Livezey, Dougherty, Gu, Berke, and
  Bouchard]{sachdeva2021improved}
Pratik~S Sachdeva, Jesse~A Livezey, Maximilian~E Dougherty, Bon-Mi Gu, Joshua~D
  Berke, and Kristofer~E Bouchard.
\newblock Improved inference in coupling, encoding, and decoding models and its
  consequence for neuroscientific interpretation.
\newblock \emph{Journal of Neuroscience Methods}, page 109195, 2021.

\bibitem[Mathuriya et~al.(2018)Mathuriya, Bard, Mendygral, Meadows, Arnemann,
  Shao, He, K{\"a}rn{\"a}, Moise, Pennycook, et~al.]{mathuriya2018cosmoflow}
Amrita Mathuriya, Deborah Bard, Peter Mendygral, Lawrence Meadows, James
  Arnemann, Lei Shao, Siyu He, Tuomas K{\"a}rn{\"a}, Diana Moise, Simon~J
  Pennycook, et~al.
\newblock Cosmoflow: using deep learning to learn the universe at scale.
\newblock In \emph{SC18: International Conference for High Performance
  Computing, Networking, Storage and Analysis}, pages 819--829. IEEE, 2018.

\bibitem[Kim et~al.(2019)Kim, Kim, Lee, Yoon, Kahou, Kashinath, and
  Prabhat]{kim2019deep}
Sookyung Kim, Hyojin Kim, Joonseok Lee, Sangwoong Yoon, Samira~Ebrahimi Kahou,
  Karthik Kashinath, and Mr~Prabhat.
\newblock Deep-hurricane-tracker: Tracking and forecasting extreme climate
  events.
\newblock In \emph{2019 IEEE Winter Conference on Applications of Computer
  Vision (WACV)}, pages 1761--1769. IEEE, 2019.

\bibitem[Livezey and Glaser(2021)]{livezey2021deep}
Jesse~A Livezey and Joshua~I Glaser.
\newblock Deep learning approaches for neural decoding across architectures and
  recording modalities.
\newblock \emph{Briefings in Bioinformatics}, 22\penalty0 (2):\penalty0
  1577--1591, 2021.

\bibitem[Kuhn(1955)]{kuhn1955hungarian}
Harold~W Kuhn.
\newblock The hungarian method for the assignment problem.
\newblock \emph{Naval research logistics quarterly}, 2\penalty0 (1-2):\penalty0
  83--97, 1955.

\bibitem[Gebru et~al.(2018)Gebru, Morgenstern, Vecchione, Vaughan, Wallach,
  Daum{\'e}~III, and Crawford]{gebru2018datasheets}
Timnit Gebru, Jamie Morgenstern, Briana Vecchione, Jennifer~Wortman Vaughan,
  Hanna Wallach, Hal Daum{\'e}~III, and Kate Crawford.
\newblock Datasheets for datasets.
\newblock \emph{arXiv preprint arXiv:1803.09010}, 2018.

\end{thebibliography}
\vfill
\pagebreak


\end{document}